\documentclass{article}

\usepackage{microtype}
\usepackage{graphicx}
\usepackage{subfigure}
\usepackage{booktabs} %

\usepackage{hyperref}
\hypersetup{
    colorlinks=true,
    linkcolor={blue!50!black},
    citecolor={blue!50!black},
    urlcolor={blue!80!black}
}

\usepackage{times}
\usepackage[utf8]{inputenc}
\usepackage{amsmath}
\usepackage{amssymb}
\usepackage{amsthm}
\usepackage{amsfonts}
\usepackage{bm}
\usepackage{pythonhighlight}
\usepackage{url}

\usepackage{listings}

\definecolor{codegreen}{rgb}{0,0.6,0}
\definecolor{codegray}{rgb}{0.5,0.5,0.5}
\definecolor{codepurple}{rgb}{0.58,0,0.82}
\definecolor{backcolour}{rgb}{0.99,0.99,0.99}

\lstdefinestyle{mystyle}{
    backgroundcolor=\color{backcolour},
    commentstyle=\color{codegreen},
    keywordstyle=\color{magenta},
    numberstyle=\tiny\color{codegray},
    stringstyle=\color{codepurple},
    basicstyle=\ttfamily\footnotesize,
    breakatwhitespace=false,
    breaklines=true,
    captionpos=b,
    keepspaces=true,
    numbers=left,
    numbersep=5pt,
    showspaces=false,
    showstringspaces=false,
    showtabs=false,
    tabsize=2
}

\usepackage[accepted]{mlsys2022}

\usepackage{amsmath,amsfonts,bm}

\def\eqref#1{equation~\ref{#1}}

\DeclareMathAlphabet{\mathsfit}{\encodingdefault}{\sfdefault}{m}{sl}
\SetMathAlphabet{\mathsfit}{bold}{\encodingdefault}{\sfdefault}{bx}{n}

\def\gT{{\mathcal{T}}}

\newcommand{\V}[1]{\bm{#1}}

\newcommand{\eq}[1]{\begin{align}#1\end{align}}

\newcommand{\brck}[1]{\left(#1\right)}

\newcommand{\brcksq}[1]{\left[#1\right]}
\newcommand{\brckcur}[1]{\left\{#1\right\}}

\def\ourname{WarpDrive}
\def\TagEnv{Tag}

\newcommand{\classname}[1]{\texttt{#1}}

\newcommand{\funcname}[1]{\textit{#1}}

\def\DataManager{\classname{DataManager}}
\def\FunctionManager{\classname{FunctionManager}}
\def\Sampler{\classname{Sampler}}
\def\EnvironmentReset{\classname{EnvironmentReset}}
\def\Environment{\classname{Environment}}
\def\EnvWrapper{\classname{EnvWrapper}}
\def\Trainer{\classname{Trainer}}
\def\EnvironmentCPUvsGPU{\classname{\EnvironmentCPUvsGPU}}

\def\nagents{N}

\def\pol{\pi}

\def\idxi{i}

\def\ac{a}
\def\vac{\V{\ac}}
\def\AC{A}

\def\st{s}
\def\vst{\V{\st}}
\def\ST{S}

\def\rew{r}
\def\vrew{\V{\rew}}

\def\df{\gamma}

\def\rlt{\xi}

\def\eplen{T}

\def\trans{\gT}

\mlsystitlerunning{
    \ourname{}: Extremely Fast End-to-End Deep Multi-Agent Reinforcement Learning on a GPU
}

\begin{document}

\twocolumn[
\mlsystitle{
    \ourname{}: Extremely Fast End-to-End Deep Multi-Agent Reinforcement Learning on a GPU
}
\mlsyssetsymbol{equal}{*}
\begin{mlsysauthorlist}
\mlsysauthor{Tian Lan}{equal,sfdc}
\mlsysauthor{Sunil Srinivasa}{equal,sfdc}
\mlsysauthor{Huan Wang}{sfdc}
\mlsysauthor{Stephan Zheng}{sfdc}
\end{mlsysauthorlist}
\mlsysaffiliation{sfdc}{Salesforce Research, Palo Alto, CA, USA}
\mlsyscorrespondingauthor{Stephan Zheng}{stephan.zheng@salesforce.com}
\mlsyskeywords{Machine Learning, Reinforcement Learning, Multi-Agent System, GPU, MLSys}
\begin{abstract}
    Deep reinforcement learning (RL) is a powerful framework to train decision-making models in complex environments. 
    However, RL can be slow as it requires repeated interaction with a simulation of the environment.
    In particular, there are key system engineering bottlenecks when using RL in complex environments that feature multiple agents with high-dimensional state, observation, or action spaces.
    We present \ourname{}, a flexible, lightweight, and easy-to-use open-source RL framework that implements end-to-end deep multi-agent RL on a single GPU (Graphics Processing Unit), built on PyCUDA and PyTorch.
    Using the extreme parallelization capability of GPUs, \ourname{} enables orders-of-magnitude faster RL compared to common implementations that blend CPU simulations and GPU models.
    Our design runs simulations and the agents in each simulation in parallel.
    It eliminates data copying between CPU and GPU.
    It also uses a single simulation data store on the GPU that is safely updated in-place.
    \ourname{} provides a lightweight Python interface and flexible environment wrappers that are easy to use and extend.
    Together, this allows the user to easily run thousands of concurrent multi-agent simulations and train on extremely large batches of experience.
    Through extensive experiments, we verify that \ourname{} provides high-throughput and scales almost linearly to many agents and parallel environments.
    For example, \ourname{} yields 2.9 million environment steps/second with 2000 environments and 1000 agents (at least $100\times$ higher throughput compared to a CPU implementation) in a benchmark \TagEnv{} simulation.
    As such, \ourname{} is a fast and extensible multi-agent RL platform to significantly accelerate research and development.
\end{abstract}
\vskip 0.3in
]

\printAffiliationsAndNotice{\mlsysEqualContribution} %

\let\svthefootnote\thefootnote
\newcommand\freefootnote[1]{%
  \let\thefootnote\relax%
  \footnotetext{#1}%
  \let\thefootnote\svthefootnote%
}

\section{Introduction}\label{sec:introduction}
Deep reinforcement learning (RL) is a powerful framework to train AI agents.
RL agents have beaten humans at several strategy games \cite{OpenAI_dota,vinyals2019grandmaster},
trained robotic arms \cite{gu2017deep},
and have been used to design economic policies \cite{zheng2021ai,trott2021building}.

However, it remains challenging to apply RL in complex simulations that feature multiple agents or high-dimensional state, observation, or action spaces, for example.
In particular, multi-agent systems are a frontier for RL research and applications, especially those with (many) interacting agents,
and are relevant to economics, dialogue agents, robotics, and many other fields.
However, there are still many engineering and scientific challenges to the use of RL.

A central challenge is that RL experiments can take days or even weeks, especially with a large number of agents.
The main reason is that the online RL-loop repeatedly runs simulation and trains agents.
Here, the number of repetitions required can grow exponentially with the complexity of the learning problem.
This is most salient in the \emph{model-free} setting, where RL agents train with zero initial knowledge about the simulation or task at hand.
This can lead to prohibitively long wall-clock training time because current deep RL implementations often combine CPU-based simulations with GPU neural network models.
RL can be especially inefficient in the multi-agent setting, as CPUs have limited potential to parallelize computations across many agents and simulations, while CPU-GPU data transfer can be slow.

Several recent works have built domain-specific, GPU or TPU-based RL solutions, for Atari \citep{dalton2020accelerating}, or learning robotic control in 3-D rigid-body simulations \citep{petrenko2021megaverse,freeman2021brax,makoviychuk2021isaac}.
These frameworks have mostly focused on (specific) single-agent problems, and are challenging to extend to \emph{multi-agent} RL. 
For example, Brax builds on JAX \cite{jax2018github} and functional programming (FP) principles.
While FP enables easy parallelization, it is challenging to use FP to build efficient multi-agent simulations, which may require manipulating complex multi-agent state representations (including mutable collections and hash tables) and graph-based or branch divergent logic to describe interactions betwee agents.
These issues become prohibitive especially with simulations with a large number of agents.

\paragraph{Creating Fast RL Pipelines with \ourname{}.}
We built \ourname{}%
\footnote{
The name \emph{\ourname{}} is inspired by the science fiction concept of a fictional superluminal spacecraft propulsion system. Moreover, at the time of writing, a \emph{warp} is a group of 32 threads that are executing at the same time in (certain) GPUs.}%
, an open-source framework to build extremely fast deep (multi-agent) RL pipelines.
\ourname{} runs the full RL workflow end-to-end on a single GPU, using a single store of data for simulation roll-outs, inference, and training.
This minimizes costly communication and copying, and significantly increases simulation sampling and learning rates.
\ourname{} also runs simulations and the agents in each simulation in tandem, capitalizing on the parallelization capabilities of GPUs.
Taken together, these design choices enable running thousands of concurrent simulations, each containing thousands of agents, and training on extremely large batches of experience.
Our benchmarks show that \ourname{} achieves orders-of-magnitude faster RL compared to common implementations that blend CPU simulations and GPU models.
For example, \ourname{} yields at least $100\times$ higher throughput with 2000 simulations and 1000 agents in a \TagEnv{} simulation (see Section \ref{sec:benchmark environments}).

\ourname{} builds on CUDA (Compute Unified Device Architecture), a popular platform and programming model that allows users to run programs (called \emph{kernels}) on (CUDA-enabled) GPU hardware.
This enables users to use the full feature set of CUDA programming, including the GPU's parallel computational elements, making it convenient to implement even complex  multi-agent simulations.
\ourname{} seamlessly integrates with any CUDA C-based simulation that has a \emph{gym}-style API~\cite{1606.01540} through a light-weight environment wrapper that executes the \emph{step} on the GPU.
It also provides a PyTorch-based trainer and training utilities that implement end-to-end RL training on the GPU.
As such, \ourname{} is flexible and easy to use and extend, and allows users to create and extend custom RL pipelines that maximize the utility of GPUs.

\begin{figure}[ht!]
\centering
    \includegraphics[width=0.618\linewidth]{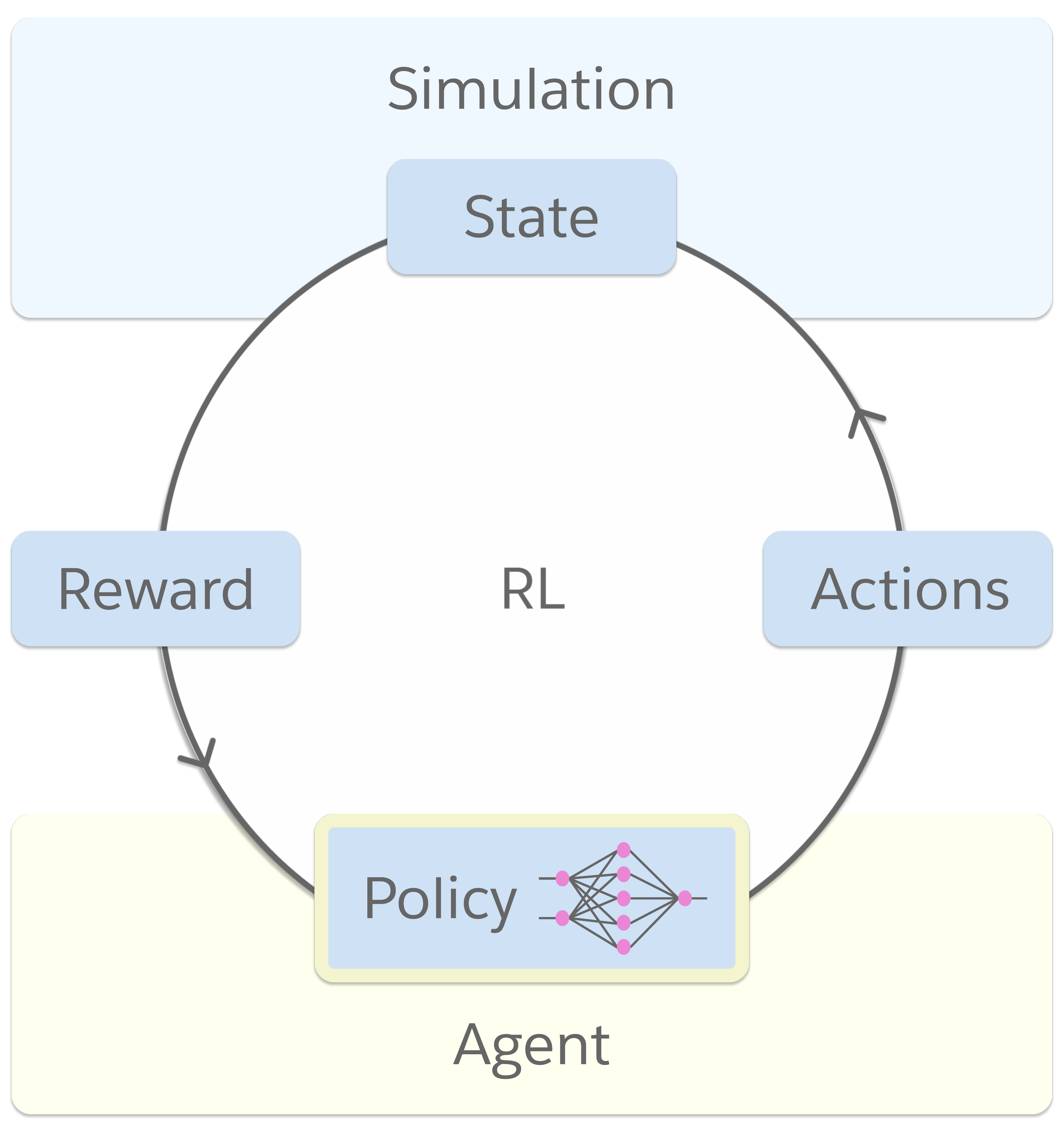}
    \caption{
    \textbf{The reinforcement learning loop for a single agent interacting with a simulation environment.}
    The agent repeatedly observes a state, receives a reward, and samples actions using its policy model. Given the agent's action, the simulation environment transitions to the next state. This structure generalizes to multiple agents interacting with the simulation, each agent being part of the environment from the point of view of the other agents.}
    \label{fig:rl-loop}
\end{figure}

\section{The RL Problem}\label{sec:The Reinforcement Learning Problem}
To set context, we summarize the RL problem \citep{sutton_reinforcement_2018}.
RL is formalized as a Markov Decision Process (MDP) and extensions thereof.
A \emph{Markov Game} is a multi-agent extension of the MDP that formally describes a system of agents, indexed by $\idxi = 1, \ldots, \nagents$ that interact with a (simulation) environment \citep{littman_markov_1994}.
The environment is further defined by a state space $\ST$, action space $\AC$, reward function $\rew$, environment dynamics $\trans\brck{\vst_{t+1} | \vst_{t}, \vac_{t}}$, and a discount factor $\df$.
Boldface quantities denote vectors over agents, e.g., $\vst = \brck{\st_1, \ldots, \st_\nagents}$.
This process is shown for a single agent in Figure \ref{fig:rl-loop}.
Each RL agent uses a \emph{policy model} $\pol_\idxi\brck{\ac_i | \st_i}$ to sample actions to execute.
Given the actions, the dynamics $\trans$ move the environment forward.
A \emph{roll-out} is a sequence of transitions $\rlt = \brckcur{\brck{\vst_t, \vac_t, \vrew_t}}_{t=0, \ldots, \eplen}$, representing the \emph{experience} of the agents in the simulation.
Given roll-outs, the goal of RL is to optimize the policies $\V{\pol} = \brck{\pol_1, \ldots, \pol_\nagents}$, each aiming to maximize its discounted expected reward:
\eq{
\pol_\idxi^* = \arg \max_{\pol_\idxi} \mathbb{E}_{\V{\pol}, \trans}\brcksq{\sum_{t=0}^\eplen \df^t \rew_{\idxi,t}}.
}
We focus on \emph{model-free}, \emph{on-policy} RL.
This means that the agents do not explicitly learn a parametric ``world model'' of $\trans$ and use the policy $\pol_\idxi$ for both exploration (collecting unseen experience) \emph{and} exploitation (executing ``optimal'' behavior).
This approach has shown its potential by yielding superhuman performance in games \citep{silver2017mastering,vinyals2019grandmaster}.
However, a downside of model-free RL is that it often requires a significant amount of roll-out data, especially when applying RL to complex problems.
As such, it is crucial to build high-throughput RL systems with fast RL-loops, as in Figure \ref{fig:rl-loop}.

\section{Distributed RL Systems}
\begin{figure}[hbt]
\centering
    \includegraphics[width=\linewidth]{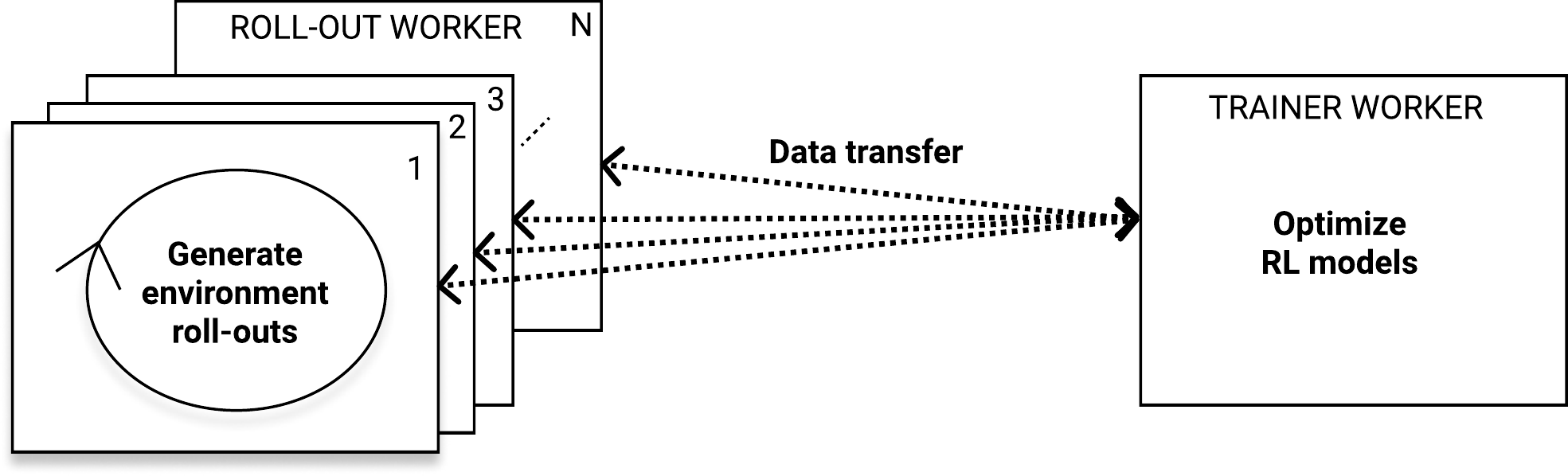}
    \caption{
    \textbf{A typical distributed RL architecture.}
    There are several roll-out workers that repeatedly generate environment roll-outs in parallel.
    The data is transferred to the trainer worker for performing policy optimization, and the updated model parameters or actions are transferred back to the roll-out workers.
    }
    \label{fig:distributed_rl}
\end{figure}

Distributed computing is a popular approach to accelerate and scale up RL systems.
Distributed RL architectures typically comprise a large number of \emph{roll-out} and \emph{trainer} workers operating in tandem (see Figure \ref{fig:distributed_rl}).
The roll-out workers repeatedly step through the environment to generate roll-outs in parallel, using the actions sampled from the policy models on the roll-out workers \cite{tian2017elf, espeholt2018impala, hoffman2020acme, pretorius2021mava} or provided by the trainer worker \cite{espeholt2020seed}.
Roll-out workers typically use CPU machines, and sometimes, GPU machines for richer environments.
Trainer workers gather the roll-out data (asynchronously) from the roll-out workers and optimize policies on CPU or GPU machines.
While these architectures are highly scalable, they have several shortcomings.

\paragraph{Expensive Communication.}
There is repeated data transfer between roll-out and trainer workers, e.g., experience from the roll-out workers to the trainer worker, and model parameters or actions from the trainer worker back to the roll-out workers.
In particular, when the environment's observation space is large and/or when the number of roll-out workers is large, the data transfer becomes very expensive.

\paragraph{Poor Utilization.}
The roll-out and trainer workers run different types of tasks with different compute requirements. 
This can lead to inefficient resource utilization. 
Calibrating the optimal ratio of worker and/or node types can be tedious.

\paragraph{Slow Simulation.}
In the context of multi-agent simulations, especially with a large number of agents, running the environment step itself can become the bottleneck, since observations, rewards, and other information needs to be computed for multiple agents.
While it's often possible to parallelize operations across agents, the roll-out time only increases with increasing the number of agents.

\paragraph{Heavy Hardware Requirements.}
Complex simulations, e.g., with multiple agents, often need a lot of compute power.
This often requires setting up a large (cluster of) node(s) with multiple processors, which can be non-trivial.

\begin{figure*}[t]
\centering
    \includegraphics[width=0.9 \linewidth]{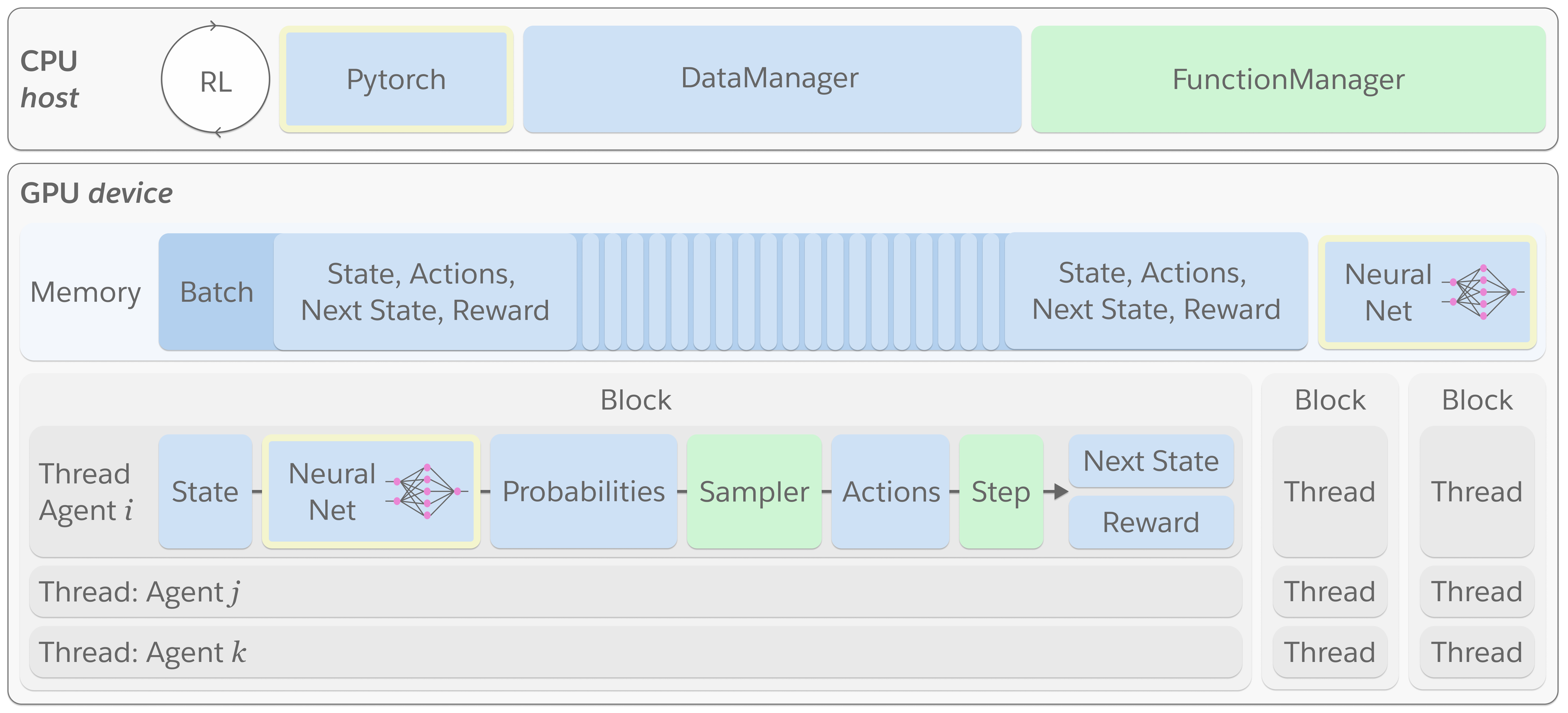}
    \caption{
    \textbf{Overview of \ourname{}'s layout of computational and data structures.}
    GPU computations are organized into blocks, each block has multiple threads.
    In this example simulation implementation, each block has a simulation environment, each thread simulates an agent.
    Blocks can access a shared GPU memory that stores simulation data and neural network policy models.
    A \DataManager{} and \FunctionManager{} enable defining RL GPU-workflows in Python.
    }
    \label{fig:warpdrive-framework-overview}
\end{figure*}

\begin{figure*}[t]
\centering
    \includegraphics[width=0.9\linewidth]{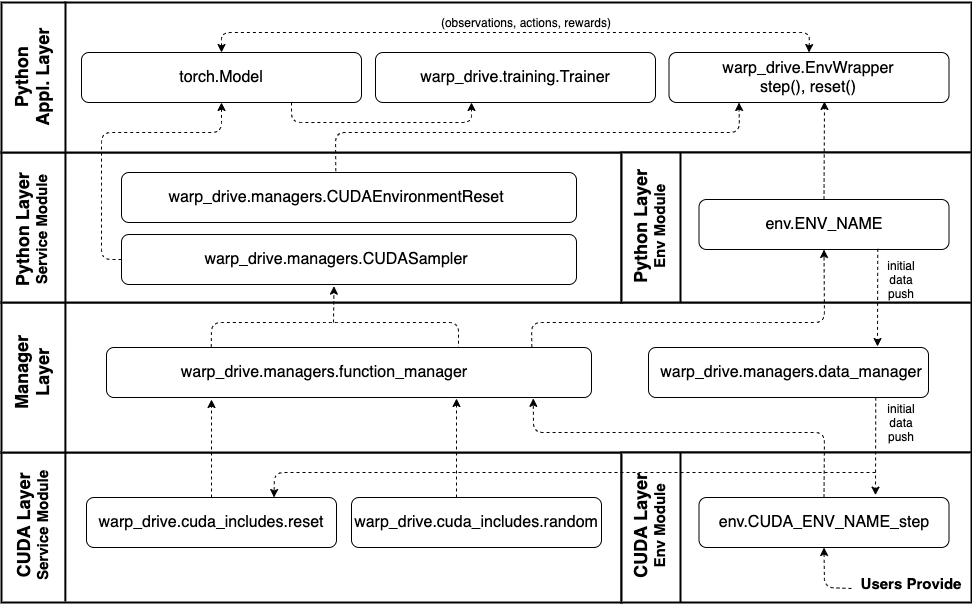}
    \caption{
    \textbf{\ourname{} code structure diagram.}
    The bottom-up overview of the main layers of \ourname{} and their relationships. Each layer or component is well separated and incrementally executable, so the user can easily create and extend custom RL pipelines.
    }
    \label{fig:warpdrive-code-structure}
\end{figure*}
\section{Accelerating RL with \ourname{}}\label{sec:Accelerating RL with WarpDrive}
\ourname{} addresses the above challenges by enabling RL workflows on a single GPU.
\ourname{} provides a framework and quality-of-life tools to implement fast and flexible multi-agent RL systems.

We emphasize that \ourname{} is \emph{complementary} to other RL systems, such as distributed RL.
Future work could implement hybrid distributed versions of \ourname{}.

We now discuss the design principles and key benefits, followed by a bottom-up overview of the design and components of \ourname{}, and describe how our design choices enable extremely fast end-to-end RL.

\subsection{Design Principles}
 \ourname{} is built following these design principles:

\begin{enumerate}
    \item Run the full end-to-end RL workflow on a GPU, including roll-out generation and training.
    \item Update data in-place to eliminate data communication.
    \item Maximally parallelize, e.g., environments and agents.
    \item Stay modular and flexible to easily accommodate using new environments, models and training algorithms. 
    \item Each part of the RL pipeline is separately testable.
    \item Use imperative and stateful code to build complex multi-agent simulation logic with interacting agents.
    \item Retain low-level control over thread mapping to environment replicas and agents for expressiveness.
    \item Maximize quality-of-life by building easy-to-use APIs and utilities for common RL pipeline components.
\end{enumerate}

\subsection{Key Benefits}
Together, these design choices enable fast end-to-end RL while fostering a scalable development ecosystem. 
That is:
\begin{enumerate}
    \item \textbf{Minimal Communication.} There is only a one-time data transfer between the CPU and the GPU (after the first \funcname{reset}), and no required communication thereafter.
    During all the subsequent \funcname{step} and \funcname{reset} calls, the data arrays are stored exclusively on the GPU and modified in-place.
    All the data on the GPU in accessed and modified in-place, so there is no data copying cost.

    \item \textbf{Extremely Fast Simulation.}
    Because each agent only uses a single \emph{thread} on the GPU, we can simulate millions of agents and/or environments, making this paradigm extremely parallelized and efficient.

    \item \textbf{Runs on a Single GPU.}
    Our current release requires only a single GPU and does not require communication between multiple GPU devices.
    It is an open direction to explore efficient multi-device RL systems.

    \item \textbf{Flexible Development Platform.} 
    The user owns the precise scheduling of each thread and has granular control over how the multi-agent logic is expressed. 
    \ourname{} also provides RL components that are lightweight and well separated. 
    It is easy to create and extend custom RL pipelines.
\end{enumerate}

\subsection{CUDA and GPU Structure}
Figure \ref{fig:warpdrive-framework-overview} illustrates the fundamental architecture design of \ourname{}.
Following the CUDA convention, the CPU is referred to as the \emph{host} and the GPU as the \emph{device}.
Running any CUDA program involves three main steps:
\begin{enumerate}
    \item \emph{Host-to-device transfer} (\emph{push}): Copying input data from the host to device memory, e.g., at the start.
    \item \emph{Invoke CUDA kernels} (\emph{execute}): Loading CUDA functions to run and caching data on the GPU for speed.
    \item \emph{Device-to-host transfer} (\emph{pull}): Copying data back from the device to host memory, e.g., once training finishes.
\end{enumerate}
Following this paradigm, \ourname{} implements a \DataManager{} and a \FunctionManager{}: two key Python classes (residing on the CPU) to facilitate all host-device communication and interactions that are relevant to RL.
The \DataManager{} handles all host-device data transfers (\emph{push} and \emph{pull}).
The \FunctionManager{} allows the user to invoke CUDA programs (or compute kernels) from the CPU and execute them on the GPU. 
These Manager classes provide simple APIs to build high-level Python applications in \ourname{}.

A key feature of GPUs is that they can run many computational \emph{threads} in parallel.
Threads are organized into \emph{thread blocks}.
Additionally, multiple thread blocks are organized into a \emph{grid} structure.
A CUDA kernel can access and define parallel computations for these threads.
In \ourname{}, each block is designed to contain an environment replica 
and each thread simulates an agent. Blocks can access a shared GPU memory that stores mini-batches of simulations data and neural network models.

\subsection{Software Layers and Components of \ourname{}}

Figure \ref{fig:warpdrive-code-structure} provides a visual overview of module structures and their relationships in \ourname{}.
At a high level, \ourname{} consists of four layers.

\paragraph{CUDA Layer.}
The \emph{CUDA layer} executes the CUDA kernel for RL step and reset.
To execute this design, \ourname{} includes two major CUDA modules:

\begin{itemize}
    \item The \emph{CUDA Service} contains the CUDA C kernel version of the environment \emph{reset} to reset individual environment replicas at the block level, and the kernel of \emph{sampler} to sample actions at the thread level where each thread is handling one agent. 
    This is the shared library for any environment.
    \item The \emph{CUDA Environment} has the CUDA C kernel of the environment \emph{step} and is separate from other \ourname{} CUDA Services. 
    \ourname{} provides several default environments and the environment loader to load custom environments provided by the user. 
    Details on how to implement custom environments are in Section \ref{sec:custom-env}.
\end{itemize}

\paragraph{Manager Layer.}
The \emph{manager layer} provides a \DataManager{} and \FunctionManager{} to communicate with the CUDA layer.
Using these managers, \ourname{} provides Python classes to host and manage the corresponding CUDA back-end and support the high-level applications built on \ourname{}.

\paragraph{Python Layer.}
These Python classes serve as fundamental classes for the Python applications running in \ourname{}, and communicate directly with the CUDA kernels. In particular, there are two major modules: 
\begin{itemize}
  \item \emph{Python Service} contains the \Sampler{} class that directly controls the CUDA \emph{sampler} kernel to sample agent actions at the thread level and maintains the action-data on the GPU. It also provides
  the \EnvironmentReset{} class that controls the CUDA \emph{reset} kernel to reset each individual environment replica in-place and independently.
  \item \emph{Python Environment} provides
  the \Environment{} class to control the CUDA \emph{step} kernel. 
\end{itemize}

\paragraph{Application Layer and Quality-of-Life Tools.}
\ourname{} provides a full development and training ecosystem for multi-agent RL on a GPU.
The \emph{application layer} supports a \emph{gym}-style interface and provides pre-built RL modules and training utilities.
\ourname{} provides several tools to simplify developing and running simulations on a GPU:
\begin{itemize}
    \item A light-weight wrapper class \EnvWrapper{} that works with the Python Service and Environment modules to automatically build \emph{gym}-style environment objects and run them on the GPU. 
    \item A \Trainer{} class, \emph{training utilities}, and \emph{example scripts} that enable end-to-end RL on a GPU in a few lines of code and easy customization of the process.
\end{itemize}

\subsection{The \ourname{} RL Workflow}
Using \ourname{}, a typical RL workflow for gathering roll-outs and training on the GPU involves the following steps:
\begin{enumerate}
    \item \textbf{One-Time Data Copy.} Copy over all the data from the host to the device only once after the environment object is initialized and reset.
    The \DataManager{} provides API methods to perform this push operation.
    The data copied from the host to the device may include environment configuration parameters, data arrays created at the end of the very first reset, as well as placeholders for the observations, actions, rewards and ``done'' flags.
    The \DataManager{} also helps maintain a copy of the variables that need to be re-initialized at every reset.
    After this point, there is no further data push from the host to the device.

    \item \textbf{Call GPU Kernels from the CPU.} 
    The \FunctionManager{} provides API methods to initialize and invoke the CUDA C kernel functions required for performing the environment step, generating observations, and computing rewards from the host node.
    These functions execute only on the device, and all the data arrays are modified in-place.
    Data may be pulled by the host from time to time for visualization or analysis purposes, but all the data can essentially reside on the GPU only during training.

    \item \textbf{Block-parallelized Environment Runs.} Within the GPU, we execute several replicas of the environment in parallel.
    Each environment runs on a separate thread block.
    Because a typical GPU has thousands of blocks, we can execute thousands of environments in parallel on just a single GPU.

    \item \textbf{Thread-parallelized Multi-agent Steps.} Within each environment (running in its own block), each agent in the environment can also execute its own logic on an agent-exclusive thread.
    Figure \ref{fig:warpdrive-framework-overview} shows an example in which agents $i$, $j$ and $k$ operate in parallel on individual threads $i$, $j$ and $k$, respectively.
    This becomes very useful in the context of multi-agent RL, since we can fully parallelize the agents' operations during the environment step, thus the simulation time complexity remains constant even as the number of agents increases (up to the number of available threads).

    \item \textbf{Automatic Environment Resetting.} Any environment may reach a terminal state and be ``done''.
    \ourname{} provides an \classname{EnvironmentReset} class designed to automatically identify and reset those environments that are done.
    At this point, those environments are also reset and given (new) initial data.
    For example, they may use the initial data arrays that were copied over at the initial reset.

    \item \textbf{Thread-parallelized Action Sampling.} 
    \ourname{} also provides a \classname{Sampler} class for sampling actions in order to step through the environment.
    Actions are sampled using the probabilities computed by the policy models.
    Our sampler runs in parallel on each agent thread, and runs about $4\times$ faster than equivalent PyTorch implementation.
    See Section \ref{sec:env_benchmark} for details.

    \item \textbf{PyTorch-based Multi-agent RL Training.} 
    Once roll-out data is gathered from several environments and agents into a training data batch, we can also perform end-to-end training with \ourname{}.
    The initial release of \ourname{} includes an example training script and \classname{Trainer} class which currently implements Advantage Actor Critic (A2C) \citep{mnih2016asynchronous} and Proximal Policy Optimization (PPO) \citep{schulman2017proximal}, and a fully-connected neural network policy model.
    The \Trainer{} builds on PyTorch, and calls all CUDA kernels and PyTorch operations (that run on the GPU device) from the CPU host.
    However, PyTorch can directly access and interpret the \ourname{} data batches (states, actions, rewards and done flags) stored on the device as a \classname{torch.Tensor}.
    This allows the user to compute losses and modify the model parameters, while eliminating data copying.
    Given the modular design of \ourname{}, it is straightforward to integrate existing implementations of other RL algorithms and model classes.
\end{enumerate}

\begin{figure}[t!]
\centering
\begin{lstlisting}[language=C]
__global__ void cuda_step(
    ..., // pointer arguments
    float * obs,
    int * actions,
    float * rew
    int* done
) {
    // Agent and environment indices 
    // corresponding to this GPU thread.
    const int kAgentIdx = threadIdx.x;
    const int kEnvIdx = blockIdx.x;
    // Update just the array indices
    // for `kAgentIdx` and `kEnvIdx`.
    ...
    // returns nothing.
}
\end{lstlisting}
\caption{
\textbf{A sample CUDA \funcname{step} function signature.}
The arguments to the \funcname{step} function are pointers to data arrays as well as the imperative observations, sampled actions, rewards and ``done'' flags that are manipulated in-place. For maximal parallelization, each GPU thread updates only the array-slices that correspond with its environment and agent.
}
\label{fig:cuda-step}
\end{figure}
\begin{figure}[ht!]
\centering
\begin{lstlisting}[language=python]
class Env:
    def __init__(self, **env_config):
        ...

    def reset(self):
        ...
        return obs

    def get_data_dictionary(self):
        # Specify the data that needs to be 
        # pushed to the GPU.
        data_feed = DataFeed()
        data_feed.add_data(
            name="variable_name",
            data=self.variable,
            save_copy_and_apply_at_reset
            =True,
        )
        ...
        return data_feed

    def step(self, actions):
        if self.use_cuda:
            self.cuda_step(
                # Pass the relevant data 
                # feed keys as arguments 
                # to cuda_step. 
                # Note: cuda_data_manager 
                # is created by the 
                # EnvWrapper.
                self.cuda_data_manager.
                device_data(...),
                ...
            )
        else:
            ...
            return obs, rew, done, info
    
\end{lstlisting}
\caption{
\textbf{Augmenting Python \classname{Environment}s for \ourname{}.}
To use an existing Python \classname{Environment} with \ourname{}, one needs to add two augmentations. 
First, a \funcname{get{\_}data{\_}dictionary()} method that returns a dictionary-like \classname{DataFeed} object with data arrays and parameters that should be pushed to the GPU.
Second, the \funcname{step}-function should call the \funcname{cuda\_step} with the data arrays that the CUDA C \funcname{step} function should have access to.
Given these additions, the \classname{EnvironmentWrapper} class can automatically build a CUDA C \classname{Environment} that handles other parts of the simulation pipeline, which includes a \DataManager{}, see Figure \ref{fig:env-wrapper}.
}
\label{fig:env-class-with-cuda}
\end{figure}
\begin{figure}[ht!]
\centering
\begin{lstlisting}[language=python]
# Create a wrapped environment object via 
# the EnvWrapper.
# Ensure that use_cuda is set to True in 
# order to run on the GPU.
env_wrapper = EnvWrapper(
    Env(**env_config), 
    num_envs=2000,
    use_cuda=True
)
# Agents can share policy models: this 
# dictionary maps policy model names to 
# agent ids.
policy_tag_to_agent_id_map = {
    <policy_tag>: [agent_ids]
}
# Create the trainer object.
trainer = Trainer(
    env_wrapper=env_wrapper,
    config=run_config,
    policy_tag_to_agent_id_map=\ 
    policy_tag_to_agent_id_map,
)
# Create and push obs, actions, rewards and 
# done flags placeholders to the device.
create_and_push_data_placeholders(
    env_wrapper,
    policy_tag_to_agent_id_map,
    trainer
)
# Perform training.
trainer.train()
\end{lstlisting}
\caption{
\textbf{\ourname{} provides quality-of-life tools to set up RL training pipeline in a few lines of code.}
\ourname{} provides the \EnvWrapper{}, and \Trainer{} classes, and utility functions to simplify building fast and flexible RL workflows.
}
\label{fig:env-wrapper}
\end{figure}
\subsection{Extensibility and Quality-of-Life Tools}
\label{sec:custom-env}
\ourname{}'s modular structure makes it easy to integrate custom Python RL environments and develop an equivalent CUDA C implementation that can run on a GPU.
At the core, the first version of \ourname{} uses simulations that are implemented using CUDA C (see Figure \ref{fig:cuda-step} for a sample function signature).
Implementing and testing programs in CUDA C can take longer than in Python.
A key reason is that memory and threads need to be carefully managed in CUDA C programs.
This is especially relevant when using GPUs, which feature multiple types of memory.

\paragraph{Checking Consistency.}
To ensure correctness of CUDA C simulations, an effective approach is to implement the simulation logic in Python and NumPy first and verify its logical correctness.
One can then implement the same logic and required data structures in CUDA C, and check whether the Python and CUDA C implementations yield similar results.
To facilitate this process, \ourname{} provides an \classname{EnvironmentCPUvsGPU} class to test consistency between Python and CUDA C implementations of the same \funcname{step} logic, i.e., whether the observations, actions, rewards and the ``done'' flags are the same at each step.

\paragraph{Building a CUDA-compatible Environment.}
Assuming the Python and CUDA C \funcname{step} functions are consistent, \ourname{} simplifies creating an augmented environment object that uses the CUDA C \funcname{step}.
First, the Python \classname{Env} class should be extended with a \funcname{get\_data\_dictionary()} method that defines which data should reside on the GPU, see Figure \ref{fig:env-class-with-cuda}.
The \EnvWrapper{} provided by \ourname{} will automatically build an augmented \classname{Environment} object that handles the low-level data transfer flow. 
This includes pushing all the data to the GPU after the very first reset, and providing \emph{gym}-style \funcname{step} and \funcname{reset} Python methods for running the simulation on the GPU.
Using this augmented environment enables RL on a GPU in a few lines of code, as shown in Figure \ref{fig:env-wrapper}.

\begin{figure}[ht!]
    \centering
\includegraphics[width=\linewidth]{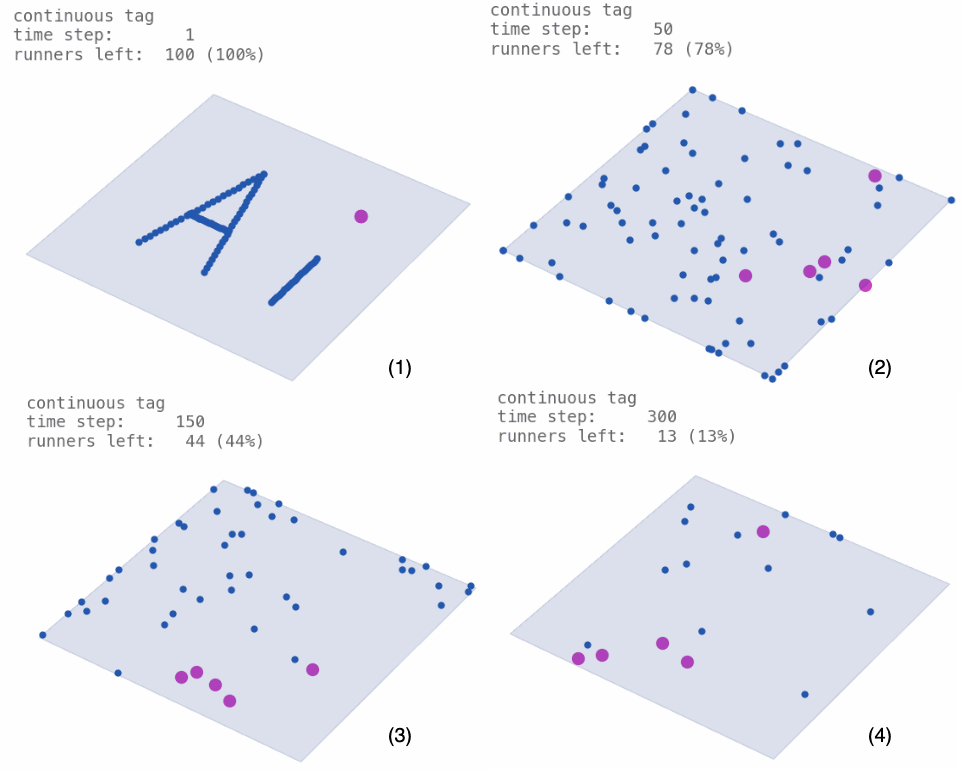}
\caption{
    \textbf{
        The \TagEnv{} environment.
    }
    This sequence of snapshots shows continuous \TagEnv{} with $5$ taggers (pink) and $100$ runners (blue) on a $20\times 20$ grid. 
    Snapshots are taken at 1) the start of the episode, 2) step 50, 3) step 150, and 4) step 300. 
    At the start, the agents are arranged in the shape of the letters ``Ai''.
    Only 13\% runners remain after 300 steps. 
    Discrete \TagEnv{} looks visually similar.
}
\label{fig:tag-viz-and-covid-19-economic-sim}
\end{figure}

\begin{figure}[ht!]
    \centering
    \includegraphics[width=\linewidth]{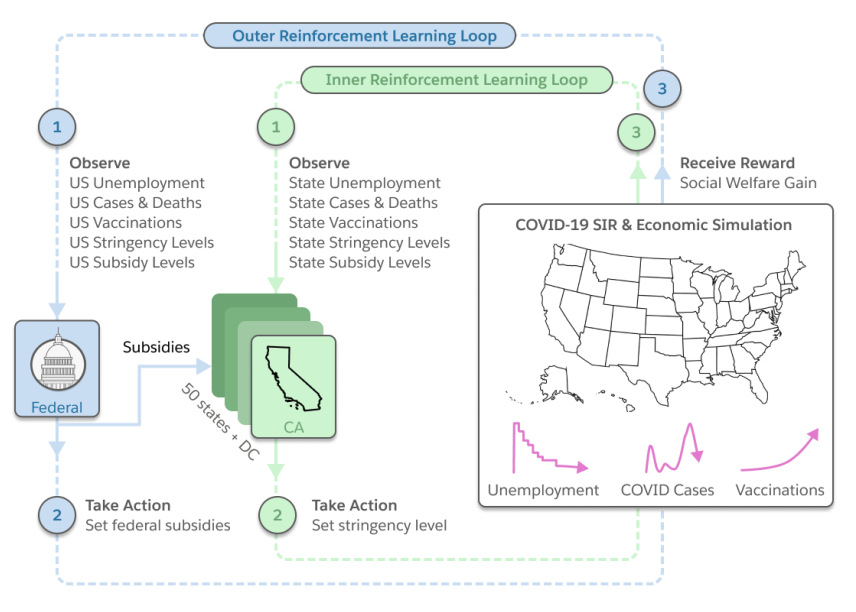}
    \caption{
        \textbf{
            COVID-19 economic simulation from \citet{trott2021building}.
        }
        This environment simulates both COVID-19 and its impact on the economy in the US, based on real-world data.
        Each agent represents a US state or the federal government. 
        Moreover, each agent optimizes its policy for its social welfare definition, a weighted sum of health (deaths) and economic outcomes (GDP) within its jurisdiction.
        A salient feature is that this is a complex, unstable \emph{two-level learning} problem, as the actions of the federal government can change the optimization problem of the states.
    }
\end{figure}

\section{Performance Benchmarks}
\label{sec:env_benchmark}
We use three environments to benchmark \ourname{}'s performance: two versions of the game of \TagEnv{} and a more complex COVID-19 economic simulation.
Our results show that the performance of \ourname{} \emph{scales linearly to thousands of environments and agents}, and yields \emph{orders of magnitude faster RL} compared to CPU implementations.

\subsection{Benchmark Environments}
\label{sec:benchmark environments}
\paragraph{The \TagEnv{} Environment.}
In \TagEnv{}, $\nagents{}_\textrm{tag} \geq 1 $ taggers work together to catch $\nagents{}_\textrm{run} \geq 1$ runners.
There are $\nagents{} = \nagents{}_\textrm{tag} + \nagents{}_\textrm{run}$ agents in total.
See Figure \ref{fig:tag-viz-and-covid-19-economic-sim} for a visualization.
Runners are tagged once a tagger gets close enough.
Each simulation episode ends after the maximum time steps, or when all runners have been tagged. 
The percentage of runners that were tagged defines how successful the taggers were.
The goal for each agent is to learn how to optimally accelerate (or brake) and turn around on the 2-D playing field.
Taggers and runners can have different skill levels; the higher the skill, the higher the maximal speed.

We use a discrete (simple) and continuous (advanced) version of \TagEnv{}.
In \emph{discrete} \TagEnv{}, agents move on a discrete 2-D grid. 
Every agent can choose to move up, down, left, right by one cell, or to not move.
In \emph{continuous} \TagEnv{}, agents move in a continuous 2-D world.
Here, every agent can accelerate, brake and/or turn around, still via a discrete set of actions, and the agents' movements follow classical mechanics.

For benchmarking, we also use two semantic variations, where agents have partial or full observations.
With partial observations, agents can only see the closest $K$ agents.

RL can optimize the tagger and runner policies.
Here, taggers are positively rewarded (e.g., $+1$) for each successful tag, so they are incentivized to tag the runners.
Once a runner is tagged, it receives a penalty (e.g., $-1$) and leaves the game.
Therefore, runners learn to avoid being tagged.
\TagEnv{} can become a complicated decision-making problem once agents are strategic (e.g., trained by RL) and as more and more taggers and runners participate.
For instance, taggers may learn cooperative strategies, e.g., taggers might learn to encircle runners.
As such, \TagEnv{} is an interesting benchmark environment for \ourname{}.

\paragraph{COVID-19 Economic Environment.}
We also show that \ourname{} scales to more complex environments, by evaluating it in a COVID-19 simulation \cite{trott2021building}.
This simulation models health and economic dynamics amidst the COVID-19 pandemic, based on real-world data. 
Figure \ref{fig:tag-viz-and-covid-19-economic-sim} shows its structure.
The simulation \funcname{step} is substantially more complex compared to \TagEnv{} and so takes a larger fraction of each iteration's run-time.

The simulation comprises 52 agents: 51 governors corresponding to each US state and Washington D.C., and another one for the the (USA) federal government.
This is a complicated \emph{two-level} multi-agent environment where the US state agents decide the stringency level of the policy response to the pandemic, while the US federal government provides subsidies to eligible individuals.
Actions taken by each agent affect its health and economic outcomes, such as deaths, unemployment, and GDP. 
In addition, the actions of the federal government can change the health-economic trade-off and optimization objective for the US states, making it a complex, unstable two-level RL problem.

\subsection{End-to-End Training Throughput}
\label{section:training throughput}
We benchmark \ourname{} by comparing performance across agents, across environment replicas, and between a 16-CPU N1 node (on GCP) and \ourname{} on an \emph{Nvidia A100} GPU.
All our benchmarks results average over $5$ repetitions.
Across the board, \ourname{} is extremely fast and yields \emph{orders of magnitude higher throughput} than when using CPU-simulations and GPU models.

\paragraph{Tag Benchmarks.}
Overall, \ourname{} achieves very fast end-to-end RL training speeds.
With 2000 discrete Tag environments and 5 agents for each environment, \ourname{} achieves \emph{1.3 million end-to-end RL training iterations per second}.
With 2000 environments and 1000 agents, it yields \emph{0.58 million} training iterations per second.
We emphasize that increasing the number of agents by $200\times$, from 5 to 1000, resulted in only $50\%$ lower throughput.
In continuous \TagEnv{}, with 2000 environments and 5 agents, \ourname{} achieves \emph{0.57 million training iterations per second},
or \emph{0.15 million training iterations per second} with 45 agents.%
\footnote{
A single A100 GPU does not have enough memory to perform end-to-end training for 2000 environments and more than 45 agents in parallel for continuous \TagEnv{}. 
}

\begin{figure}[t!]
\centering
    \includegraphics[width=0.7\linewidth]{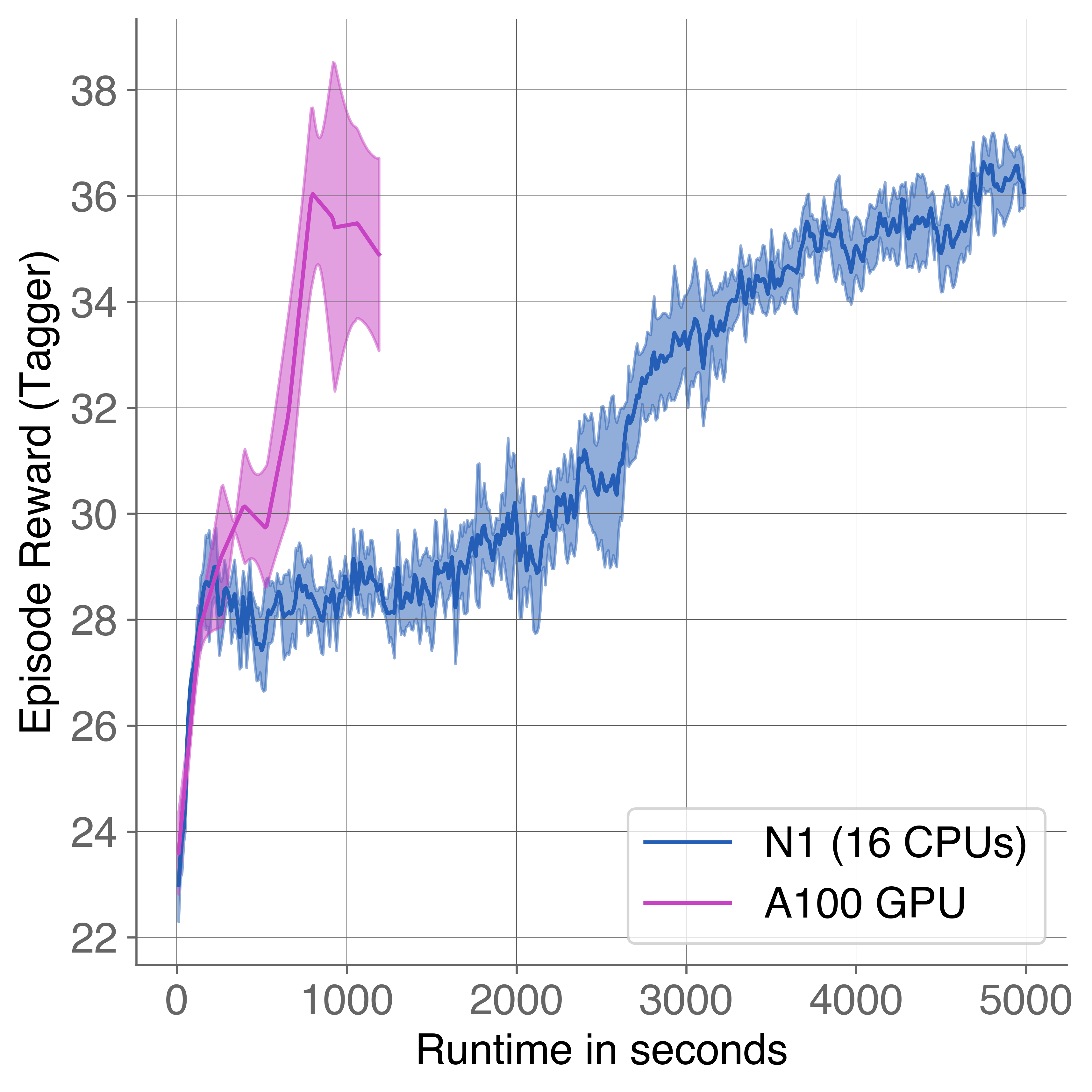}
    \caption{
    \textbf{Episode reward vs wall-clock time, continuous \TagEnv{}, 12 agents.}
    \ourname{} (GPU) reaches the same reward level $5\times$ faster than on a 16-node CPU. 
    Note that with 12 agents, CPUs have less computational overhead per environment step and can benefit from some parallelization. 
    That makes this a fairer comparison between CPU and GPU implementations than with more agents.
    }
    \label{fig:warpdrive-vs-cpu-rew_curves}
\end{figure}

Figure \ref{fig:warpdrive-vs-cpu-rew_curves} compares training speed between an N1 16-CPU node and a single A100 GPU in continuous \TagEnv{} with $10$ runners and $2$ taggers, both using $60$ environment replicas.
With the same environment and training parameters, \ourname{} on a GPU is $5\times$ faster, even with just $12$ agents.

\paragraph{COVID-19 Benchmarks.}
For the COVID-19 economic environment, \emph{\ourname{} achieves $24\times$ more steps per second} with $60$ environment replicas, compared to a $16$ CPU node.
\begin{figure}[t!]
    \centering
    \includegraphics[width=\linewidth]{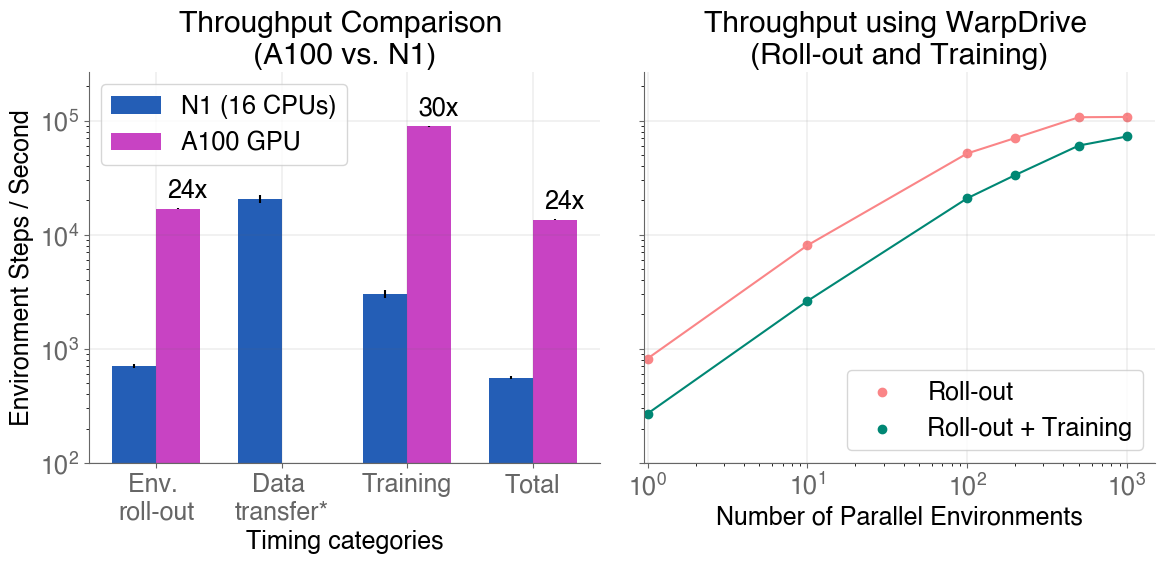}
    \caption{
        \textbf{
            \ourname{} performance in the COVID-19 economic simulation.
            Note: log-scales.
        }
        Left: $^*$: Note that there is no data transfer with \ourname{}.
        With $60$ parallel environments, WarpDrive achieves $24\times$ higher throughput over CPU-based RL training architectures (``total'').
        Moreover, both the roll-out and training phase are an order of magnitude faster than on CPU.
        Right: Environment steps per second and end-to-end training speed scale (almost) linearly with the number of environments.
    }
    \label{fig:warpdrive-covid-19-perf}    
\end{figure}
Across different timing categories (see Figure \ref{fig:warpdrive-covid-19-perf}, the performance gains comprise a $24\times$ speed-up during the environment roll-out, a zero data transfer time, and a $30\times$ speed-up for training the policy models.
Moreover, \ourname{} can scale almost linearly to 1000 parallel COVID-19 environments, resulting in even higher throughput gains.

\begin{figure}[t!]
\centering
    \includegraphics[width=\linewidth]{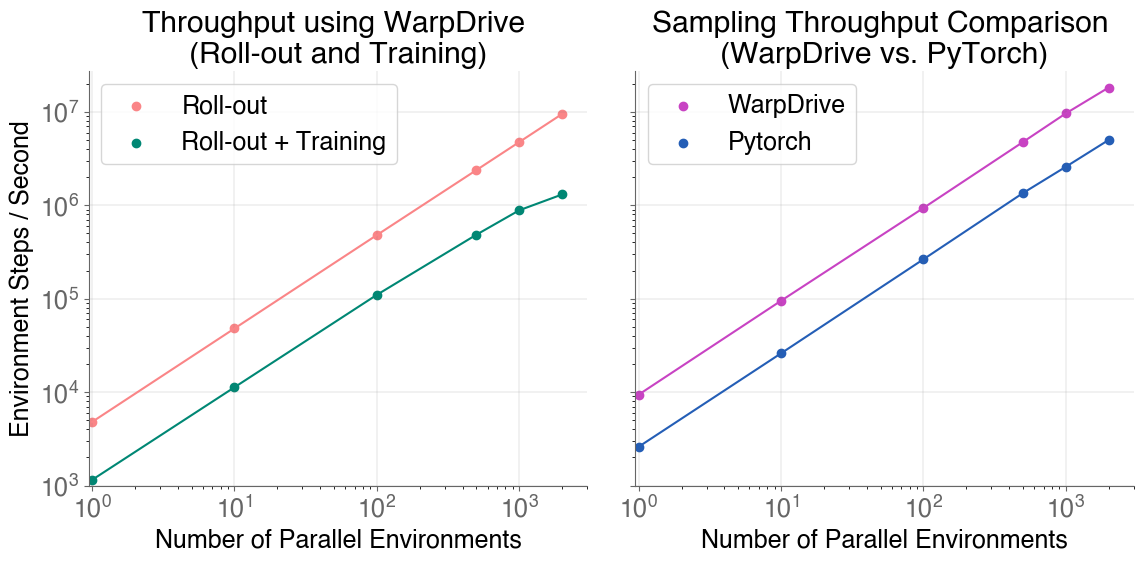}
    \caption{
    \textbf{
    \ourname{} roll-out, training, and sampling throughput in discrete \TagEnv{} with increasing number of environments with 5 agents. Note: log-log scale.
    } 
    Left: The rate of environment steps and end-to-end (roll-out + training) loops, measured in iterations per second, scales (almost) linearly.
    Right: Sampling with \ourname{} is $3.6\times$ faster than PyTorch. 
    }
    \label{fig:warpdrive-discrete-tag-perf-1}
\end{figure}

\begin{figure}[t!]
\centering
    \includegraphics[width=\linewidth]{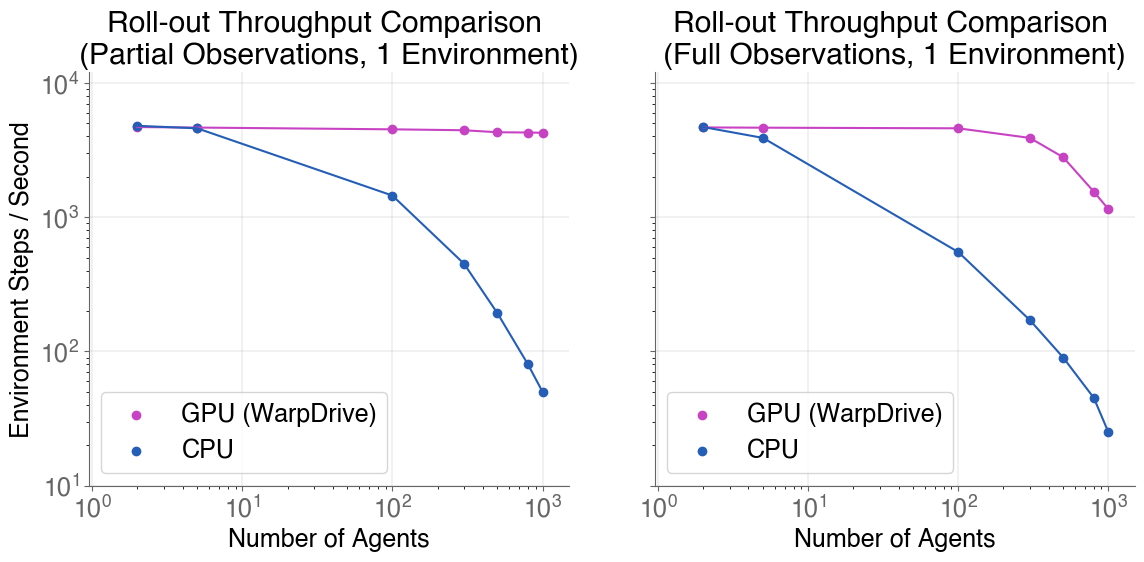}
    \caption{
    \textbf{\ourname{} roll-out throughput in discrete \TagEnv{} for increasing numbers of agents.
    Note: log-log scale.
    }
    Left: partial observations.
    Right: full observations.
    \ourname{} achieves significantly more environment steps per second, even with 1000 agents.
    Across 5 repetitions, results varied less than 3\%.
    }
    \label{fig:warpdrive-discrete-tag-perf-2}
\end{figure}

\begin{figure}[t!]
\centering
    \includegraphics[width=\linewidth]{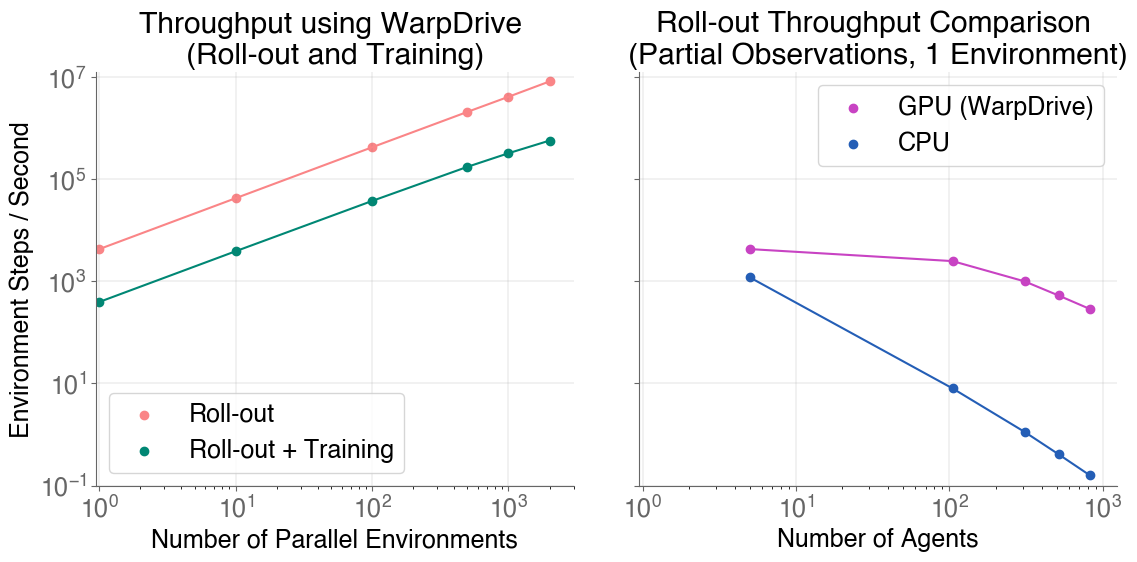}
    \caption{
    \textbf{
    \ourname{} performance in continuous \TagEnv{}.
    Note: log-log scale.
    }
    Left: The rate of environment steps and end-to-end training loops, measured in iterations per second, scale linearly with the number of environments. Each environment has 5 agents.
    Right: Using \ourname{}, the number of environment steps per second (with partial observations) is significantly higher than with CPU simulations for increasing numbers of agents.
    Across 5 repetitions, results varied less than 3\%.
    }
    \label{fig:warpdrive-continuous-tag-perf}
\end{figure}

\subsection{Scaling to Many Environments and Agents}
\ourname{} achieves nearly perfect parallelism over thousands of environments and up to one thousand agents, running on one single GPU.

Figure \ref{fig:warpdrive-discrete-tag-perf-1} (Left) shows \ourname{}'s performance in discrete \TagEnv{}.
\ourname{} \emph{scales linearly to over thousands of environments} (fixed number of agents) and yields almost perfect parallelism over environments.
For example, \ourname{} runs at \emph{9.8 million environment steps per second} with 5 agents and 2000 discrete \TagEnv{} environments.
With 1000 agents, it achieves \emph{2.9 million environment steps per second}.

Figure \ref{fig:warpdrive-discrete-tag-perf-2} shows performance per discrete \TagEnv{} environment as the number of agents grows.
For each environment replica, \emph{\ourname{} is at least $50\times$ faster} compared to a NumPy version on a single CPU, for up to 1000 agents.

Continuous \TagEnv{} is significantly faster too.
Figure \ref{fig:warpdrive-continuous-tag-perf} shows that throughput scales linearly to over thousands of environment replicas in continuous \TagEnv{}.
In particular, \ourname{} reaches \emph{8.3 million environment steps per second} with 5 agents and 2000 environments.
For each replica, \emph{\ourname{} yields at least $500\times$ more environment steps per second} compared to a single CPU, for up to 820 agents.

\subsection{Faster Sampling}
The improved performance of the \ourname{} sampler contributes to overall faster training.
In discrete \TagEnv{}, \ourname{} samples \emph{18 million actions per second per agent} with 2000 environments, independent of the number of agents (see Figure \ref{fig:warpdrive-discrete-tag-perf-1}, right). 
This is $3.6\times$ faster compared to the equivalent PyTorch operator, which yields 5 million samples per second.
In continuous \TagEnv{}, \ourname{} samples \emph{16 million actions per action category per second per agent} with 2000 environments, independent of the number of agents.

\subsection{Impact of Simulation Complexity}
The complexity of the simulation logic, as implemented in the \emph{step} and \emph{reset} function, impacts performance.
To quantify the impact of this aspect, we compared two variations of \TagEnv{}: with agents using partial observation vectors or full observation vectors.
When using partial observations, each agent can only see its $K$ nearest neighbors.
In discrete \TagEnv{}, using partial observations yields an environment step function with close to $O(\nagents{})$ time complexity, better than $O(\nagents{}^2)$.
More generally, using partial observations can enable better scaling behavior when using GPUs.
Constructing partial observations for any agent may require less information about and communication between (other) agent threads, and thus benefits more from parallelizing across agents.
However, this depends on the specific implementation of each simulation and is an important design choice.
Finally, we note that the speed gains persist under the more complex COVID-19 economic simulation, see Section \ref{section:training throughput}.
\section{Future Directions}
Future work could explore how multi-GPU setups can further improve throughput.
Furthermore, a key remaining bottleneck is to build robust CUDA simulations.
Towards making RL usable and useful, we hope \ourname{} encourages the creation of new tools to simplify simulation development in CUDA.
Being modular, we hope to extend \ourname{} and integrate other tools for building machine learning workflows on GPUs and other accelerators.  
In all, we hope that \ourname{} contributes to the democratization of high-performance RL systems and future advances in AI.

\bibliography{main}
\bibliographystyle{mlsys2022}

\appendix

\end{document}